\documentclass{article}

\usepackage{hyperref}
\usepackage{url}
\usepackage[export]{adjustbox}
\usepackage{subcaption}
\usepackage{amsmath,amsfonts,amsthm,amssymb} 
\usepackage{wrapfig}
\usepackage[normalem]{ulem}
\usepackage{amsmath}
\usepackage{todonotes}
\usepackage{natbib}
\usepackage{times}
\usepackage{graphicx} 

\usepackage{natbib}

\usepackage{algorithm}
\usepackage{algorithmic}

\usepackage{hyperref}

\usepackage[accepted]{icml2016}

\icmltitlerunning{Auxiliary Deep Generative Models}

\begin{document} 

\twocolumn[
\icmltitle{Auxiliary Deep Generative Models}

\icmlauthor{Lars Maal\o e$^1$}{larsma@dtu.dk}
\icmlauthor{Casper Kaae S\o nderby$^2$}{casperkaae@gmail.com}
\icmlauthor{S\o ren Kaae S\o nderby$^2$}{skaaesonderby@gmail.com}
\icmlauthor{Ole Winther$^{1,2}$}{olwi@dtu.dk}
\icmladdress{$^1$Department of Applied Mathematics and Computer Science, Technical University of Denmark\\
$^2$Bioinformatics Centre, Department of Biology, University of Copenhagen}

\icmlkeywords{variational inference, semi-supervised learning, deep learning, neural networks}

\vskip 0.3in
]

\begin{abstract}
Deep generative models parameterized by neural networks have recently achieved state-of-the-art performance in unsupervised and semi-supervised learning. We extend deep generative models with auxiliary variables which improves the variational approximation. The auxiliary variables leave the generative model unchanged but make the variational distribution more expressive. Inspired by the structure of the auxiliary variable we also propose a model with two stochastic layers and skip connections. Our findings suggest that more expressive and properly specified deep generative models converge faster with better results. We show state-of-the-art performance within semi-supervised learning on MNIST, SVHN and NORB datasets.
\vspace{-4mm}
\end{abstract}

\section{Introduction}
\label{introduction}
Stochastic backpropagation, deep neural networks and approximate Bayesian inference have made deep generative models practical for large scale problems \citep{Kingma13,Rezende14}, but typically they assume a mean field latent distribution where all latent variables are independent. This assumption might result in models that are incapable of capturing all dependencies in the data. In this paper we show that deep generative models with more expressive variational distributions are easier to optimize and have better performance. We increase the flexibility of the model by introducing auxiliary variables \citep{Agakov04} allowing for more complex latent distributions. We demonstrate the benefits of the increased flexibility by achieving state-of-the-art performance in the semi-supervised setting for the MNIST \citep{LeCun98}, SVHN \citep{Netzer2011} and NORB \citep{LeCun2004} datasets.

Recently there have been significant improvements within the semi-supervised classification tasks. \citet{Kingma14} introduced a deep generative model for semi-supervised learning by modeling the joint distribution over data and labels. This model is difficult to train end-to-end with more than one layer of stochastic latent variables, but coupled with a pretrained feature extractor it performs well. Lately the Ladder network \citep{Rasmus15, Valpola14} and virtual adversarial training (VAT) \citep{Miyato15} have improved the performance further with end-to-end training.

In this paper we train deep generative models with multiple stochastic layers. The \emph{ Auxiliary Deep Generative Models (ADGM)} utilize an extra set of auxiliary latent variables to increase the flexibility of the variational distribution (cf. Sec. \ref{sec:aux}).  We also introduce a slight change to the ADGM, a 2-layered stochastic model with skip connections, the \emph{ Skip Deep Generative Model (SDGM)} (cf. Sec. \ref{sec:sdgm}). Both models are trainable end-to-end and offer state-of-the-art performance when compared to other semi-supervised methods.
%
In the paper we first introduce toy data to demonstrate that:
\vspace{-3mm}
\begin{itemize}
\setlength\itemsep{0.00em}
\item[(i)] The auxiliary variable models can fit complex latent distributions and thereby improve the variational lower bound (cf. Sec. \ref{sec:vae_toy} and \ref{sec:loglikelihood}).
\item[(ii)] By fitting a complex half-moon dataset using only six labeled data points the ADGM utilizes the data manifold for semi-supervised classification (cf. Sec. \ref{sec:halfmoons}).
\end{itemize}
\vspace{-3mm}
For the benchmark datasets we show (cf. Sec. \ref{sec:benchmarks}):
\vspace{-3mm}
\begin{itemize}
\setlength\itemsep{0.00em}
\item[(iii)] State-of-the-art results on several semi-supervised classification tasks. 
\item[(iv)] That multi-layered deep generative models for semi-supervised learning are trainable end-to-end without pre-training or feature engineering.
\end{itemize}

\section{Auxiliary deep generative models}
Recently \citet{Kingma13,Rezende14} have coupled the approach of variational inference with deep learning giving rise to powerful probabilistic models constructed by an inference neural network $q(z|x)$ and a generative neural network $p(x|z)$. This approach can be perceived as a variational equivalent to the deep auto-encoder, in which $q(z|x)$ acts as the encoder and $p(x|z)$ the decoder. However, the difference is that these models ensures efficient inference over continuous distributions in the latent space $z$ with automatic relevance determination and regularization due to the KL-divergence. Furthermore, the gradients of the variational upper bound are easily defined by backpropagation  through the network(s). To keep the computational requirements low the variational distribution $q(z|x)$ is usually chosen to be a diagonal Gaussian, limiting the expressive power of the inference model.

\begin{figure}[h!]
  \centering
  	\begin{subfigure}{.22\textwidth}
	  \centering
      \includegraphics[width=1.\textwidth]{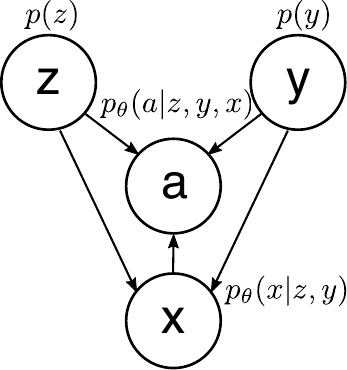}
	  \caption{Generative model $P$.}
	  \label{fig:p_model}
	\end{subfigure}
	\begin{subfigure}{.22\textwidth}
	  \centering
      \includegraphics[width=1.12\textwidth]{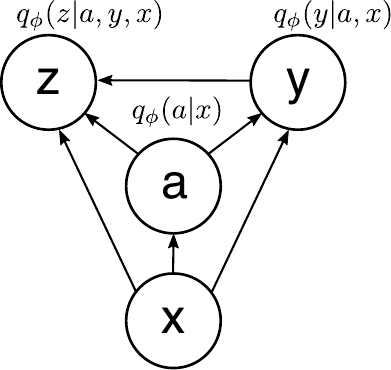}
	  \caption{Inference model $Q$.}
	  \label{fig:q_model}
	\end{subfigure}
     \vspace{-3mm}
	\caption{Probabilistic graphical model of the ADGM for semi-supervised learning. The incoming joint connections to each variable are deep neural networks with parameters $\theta$ and $\phi$.}
    \vspace{-4mm}
  \label{fig:models}
\end{figure}

In this paper we propose a variational auxiliary variable approach \cite{Agakov04} to improve the variational distribution: The generative model is extended with variables $a$ to $p(x,z,a)$ such that the original model is invariant to marginalization over $a$: $p(x,z,a)=p(a|x,z)p(x,z)$. In the variational distribution, on the other hand, $a$ is used such that marginal $q(z|x)= \int q(z|a,x)p(a|x)da$ is a general non-Gaussian distribution. This hierarchical specification allows the latent variables to be correlated through $a$, while maintaining the computational efficiency of fully factorized models (cf. Fig. \ref{fig:models}). In Sec. \ref{sec:vae_toy} we demonstrate the expressive power of the inference model by fitting a complex multimodal distribution.

\subsection{Variational auto-encoder}
The variational auto-encoder (VAE) has recently been introduced as a powerful method for unsupervised learning. Here a latent variable generative model $p_\theta(x|z)$ for data $x$ is parameterized by a deep neural network with parameters $\theta$. The parameters are inferred by maximizing the variational lower bound of the likelihood $-\sum_i {\cal U_\text{VAE}}(x_i)$ with 
\begin{align}\label{eq:vae_bound}
\log p(x) &= \log \int_z p(x,z) dz \nonumber \\
&\ge\mathbb{E}_{q_\phi(z|x)}
\left[ \log
\frac{p_\theta(x|z)p_\theta(z)}{q_\phi(z|x)} 
\right] \\
&\equiv - {\cal U}_\text{VAE}(x) \ . \nonumber
\end{align}

The inference model $q_\phi(z|x)$ is parameterized by a second deep neural network. The inference and generative parameters, $\theta$ and $\phi$, are jointly trained by optimizing Eq. \ref{eq:vae_bound} with stochastic gradient ascent, where we use the reparameterization trick for backpropagation through the Gaussian latent variables \citep{Kingma13,Rezende14}.

\subsection{Auxiliary variables} \label{sec:aux}
We propose to extend the variational distribution with auxiliary variables $a$: $q(a,z|x)=q(z|a,x)q(a|x)$ such that the marginal distribution $q(z|x)$ can fit more complicated posteriors $p(z|x)$. In order to have an unchanged generative model, $p(x|z)$, it is required that the joint mode $p(x,z,a)$ gives back the original $p(x,z)$ under marginalization over $a$, thus $p(x,z,a)=p(a|x,z)p(x,z)$. Auxiliary variables are used in the EM algorithm and Gibbs sampling and have previously been considered for variational learning by \citet{Agakov04}. Concurrent with this work \citet{Ranganath2015} have proposed to make the parameters of the variational distribution stochastic, which leads to a similar model. It is important to note that in order not to fall back to the original VAE model one has to require $p(a|x,z)\neq p(a)$, see \citet{Agakov04} and App.~\ref{app:auxiliary_variable}. The auxiliary VAE lower bound becomes 
%
%
%
\begin{align}\label{eq:avae_bound}
\log p(x) &= \log \int_a \int_z p(x, a, z) dadz \nonumber \\
&\ge \mathbb{E}_{q_\phi(a,z|x)}
\left[ \log
\frac{p_\theta(a|z,x) p_\theta(x|z)p(z)}{  q_\phi(a|x)q_\phi(z|a,x)} 
\right]\\
&\equiv - {\cal U}_{\text{AVAE}}(x)\ . \nonumber
\end{align}
with $p_\theta(a|z,x)$ and $q_\phi(a|x)$ diagonal Gaussian distributions parameterized by deep neural networks.

\subsection{Semi-supervised learning}
The main focus of this paper is to use the auxiliary approach to build semi-supervised models that learn classifiers from labeled and unlabeled data. To encompass the class information we introduce an extra latent variable $y$. The generative model $P$ is defined as $p(y)p(z)p_\theta(a|z,y,x)p_\theta(x|y,z)$ (cf. Fig. \ref{fig:p_model}):
\begin{align}
p(z) &= \mathcal{N}(z|0,\mathrm{I}),\\ 
p(y) &= \text{Cat}(y|\pi),\\
p_\theta(a|z,y,x) &= f(a;z, y, x,\theta),\\
p_\theta(x|z,y) &= f(x;z, y, \theta)\ ,
\end{align}
where $a$, $y$, $z$ are the auxiliary variable, class label, and latent features, respectively. $\text{Cat}(\cdot)$ is a multinomial distribution, where $y$ is treated as a latent variable for the unlabeled data points. In this study we only experimented with categorical labels, however the method applies to other distributions for the latent variable $y$. $f(x;z, y, \theta)$ is iid categorical or Gaussian for discrete and continuous observations $x$. $p_\theta(\cdot)$ are deep neural networks with parameters $\theta$. The inference model is defined as $q_\phi(a|x)q_\phi(z|a,y,x)q_\phi(y|a,x)$ (cf. Fig. \ref{fig:q_model}): 
\begin{align}
q_\phi(a|x) &= \mathcal{N}(a|\mu_\phi (x), \text{diag}(\sigma_\phi^2(x))),\\
q_\phi(y|a,x) &= \text{Cat}(y|\pi_\phi(a,x)), \\ \label{adgm:classifier}
q_\phi(z|a, y, x) &= \mathcal{N}(z|\mu_\phi (a,y,x), \text{diag}(\sigma_\phi^2(a,y,x)))\ .
\end{align}
In order to model Gaussian distributions $p_\theta(a|z,y,x)$, $p_\theta(x|z,y)$, $q_\phi(a|x)$ and $q_\phi(z|a, y, x)$ we define two separate outputs from the top deterministic layer in each deep neural network, $\mu_{\phi \vee \theta}(\cdot)$ and $\log\sigma^2_{\phi \vee \theta}(\cdot)$. From these outputs we are able to approximate the expectations $\mathbb{E}$ by applying the reparameterization trick.

The key point of the ADGM is that the auxiliary unit $a$ introduce a latent feature extractor to the inference model giving a richer mapping between $x$ and $y$. We can use the classifier (\ref{adgm:classifier}) to compute probabilities for unlabeled data $x_u$ being part of each class and to retrieve a cross-entropy error estimate on the labeled data $x_l$. This can be used in cohesion with the variational lower bound to define a good objective function in order to train the model end-to-end.

\subsubsection*{Variational Lower Bound}
We optimize the model by maximizing the lower bound on the likelihood (cf.~App.~\ref{app:variational_bounds} for more details). The variational lower bound on the marginal likelihood for a single \emph{labeled data point} is
\begin{align}
&\log p(x,y) = \log \int_a \int_z p(x,y,a,z)dzda \nonumber \\
&\ge \mathbb{E}_{q_\phi(a,z|x,y)}
\left[ \log
\frac{p_\theta(x,y,a,z)}{q_\phi(a,z|x,y)} 
\right] \\
&\equiv  - \mathcal{L}(x,y)\ , \nonumber
\end{align}
with $q_\phi(a,z|x,y) = q_\phi(a|x)q_\phi(z|a,y,x)$.
For unlabeled data we further introduce the variational distribution for $y$, $q_\phi(y|a,x)$:
\begin{align}
&\log p(x) = \log \int_a \int_y \int_z p(x,y,a,z) dzdyda \nonumber \\
&\ge  \mathbb{E}_{q_\phi(a,y,z|x)}
\left[ \log
\frac{p_\theta(x,y,a,z)}{q_\phi(a,y,z|x)} 
\right]  \\
&\equiv - \mathcal{U}(x) \ , \nonumber
\end{align}
with $q_\phi(a,y,z|x) = q_\phi(z|a,y,x) q_\phi(y|a,x) q_\phi(a|x)$.

The classifier (\ref{adgm:classifier}) appears in $-\mathcal{U}(x_u)$, but not in $-\mathcal{L}(x_l,y_l)$. The classification accuracy can be improved by introducing an explicit classification loss for labeled data:
\begin{align}
\mathcal{L}_l(x_l,y_l) &=  \label{eq:elbo_trick}\\
&\mathcal{L}(x_l,y_l) + \alpha \cdot \mathbb{E}_{q_\phi(a|x_l)}[-\log q_\phi(y_l|a,x_l)] \ , \nonumber
\end{align} 
where $\alpha$ is a weight between generative and discriminative learning. The $\alpha$ parameter is set to $\beta \cdot \frac{N_l+N_u}{N_l}$, where $\beta$ is a scaling constant, $N_l$ is the number of labeled data points and $N_u$ is the number of unlabeled data points. The objective function for labeled and unlabeled data is
\begin{align}
\mathcal{J} = \sum_{(x_l,y_l)} \mathcal{L}_l(x_l,y_l) + \sum_{(x_u)} \mathcal{U}(x_u) \ .  \label{eq:elbo_col}
\end{align}

\subsection{Two stochastic layers with skip connections} \label{sec:sdgm}
\citet{Kingma14} proposed a model with two stochastic layers but were unable to make it converge end-to-end and instead resorted to layer-wise training. In our preliminary analysis we also found that this model: $p_\theta(x|z_1)p_\theta(z_1|z_2,y)p(z_2)p(y)$ failed to converge when trained end-to-end. On the other hand, the auxiliary model can be made into a two-layered stochastic model by simply reversing the arrow between $a$ and $x$ in Fig.~\ref{fig:p_model}. We would expect that if the auxiliary model works well in terms of convergence and performance then this two-layered model ($a$ is now part of the generative model): $p_\theta(x|y,a,z)p_\theta(a|z,y)p(z)p(y)$ should work even better because it is a more flexible generative model. The variational distribution is unchanged: $q_\phi(z|y,x,a)q_\phi(y|a,x)q_\phi(a|x)$. We call this the \emph{Skip Deep Generative Model (SDGM)} and test it alongside the auxiliary model in the benchmarks (cf. Sec. \ref{sec:benchmarks}).

\section{Experiments}\label{sec:experimenta_details}
The SDGM and ADGM are each parameterized by 5 neural networks (NN): (1) auxiliary inference model $q_\phi(a|x)$, (2) latent inference model $q_\phi(z|a,y,x)$, (3) classification model $q_\phi(y|a,x)$, (4) generative model $p_\theta(a|\cdot)$, and (5) the generative model $p_\theta(x|\cdot)$.

The neural networks consists of $M$ fully connected hidden layers with $h_j$ denoting the output of a layer $j=1,...,M$. All hidden layers use rectified linear activation functions. To compute the approximations of the stochastic variables we place two independent output layers after $h_M$, $\mu$ and $\log \sigma^2$. In a forward-pass we are propagating the input $x$ through the neural network by
\begin{align}
h_M =& \mathtt{NN}(x)\\
\mu =& \mathtt{Linear}(h_M)\\
\log \sigma^2 =& \mathtt{Linear}(h_M) \ ,
\end{align}
with $\mathtt{Linear}$ denoting a linear activation function. We then approximate the stochastic variables by applying the reparameterization trick using the $\mu$ and $\log \sigma^2$ outputs.

In the unsupervised toy example (cf. Sec. \ref{sec:vae_toy}) we applied 3 hidden layers with ${\rm dim}(h)=20$, ${\rm dim}(a)=4$ and ${\rm dim}(z)=2$. For the semi-supervised toy example (cf. Sec. \ref{sec:halfmoons}) we used two hidden layers of ${\rm dim}(h)=100$ and ${\rm dim}(a,z)=10$.

For all the benchmark experiments (cf. Sec. \ref{sec:benchmarks}) we parameterized the deep neural networks with  two fully connected hidden layers. Each pair of hidden layers was of size ${\rm dim}(h)=500$ or ${\rm dim}(h)=1000$ with ${\rm dim}(a,z)=100$  or ${\rm dim}(a,z)=300$. The generative model was $p(y)p(z)p_\theta(a|z,y)p_\theta(x|z,y)$ for the ADGM and the SDGM had the augmented $p_\theta(x|a,z,y)$. Both have unchanged inference models (cf. Fig. \ref{fig:q_model}).

All parameters are initialized using the \citet{Glorot10} scheme. The expectation over the $a$ and $z$ variables were performed by Monte Carlo sampling using the reparameterization trick \citep{Kingma13,Rezende14} and the average over $y$ by exact enumeration so 
\begin{align}
\mathbb{E}_{q_\phi(a,y,z|x)}
&\left[f(a,x,y,z)\right] \approx \\ 
&\frac{1}{N_{\rm samp}} \sum_i^{N_{\rm samp}} \sum_y q_\phi(y|a_i,x) f(a_i,x,y,z_{yi}), \nonumber
\end{align}
with $a_i \sim q(a|x)$ and $z_{yi} \sim q(z|a,y,x)$.

For training, we have used the \textit{Adam} \citep{Kingma14a} optimization framework with a learning rate of 3e-4, exponential decay rate for the 1st and 2nd moment at $0.9$ and $0.999$, respectively. The $\beta$ constant was between $0.1$ and $2$ throughout the experiments.

The models are implemented in Python using Theano \citep{Bastien12}, Lasagne \citep{Dieleman15} and Parmesan libraries\footnote{Implementation is available in a repository named auxiliary-deep-generative-models on \url{github.com}.}. 

For the MNIST dataset we have combined the training set of $50000$ examples with the validation set of $10000$ examples. The test set remained as is. We used a batch size of $200$ with half of the batch always being the $100$ labeled samples. The labeled data are chosen randomly, but distributed evenly across classes. To speed up training, we removed the columns with a standard deviation below $0.1$ resulting in an input size of ${\rm dim}(x)=444$. Before each epoch the normalized MNIST images were binarized by sampling from a Bernoulli distribution with mean parameter set to the pixel intensities.

For the SVHN dataset we used the vectorized and cropped training set ${\rm dim}(x)=3072$ with classes from $0$ to $9$, combined with the \textit{extra} set resulting in $604388$ data points. The test set is of size $26032$. We trained on the \textit{small} NORB dataset consisting of $24300$ training samples and an equal amount of test samples distributed across 5 classes: \textit{animal}, \textit{human}, \textit{plane}, \textit{truck}, \textit{car}. We normalized all NORB images following \citet{Miyato15} using image pairs of $32\text{x}32$ resulting in a vectorized input of ${\rm dim}(x)=2048$. The labeled subsets consisted of $1000$ evenly distributed labeled samples. The batch size for SVHN was $2000$ and for NORB $200$, where half of the batch was labeled samples. To avoid the phenomenon on modeling discretized values with a real-valued estimation \citep{Uria2013}, we added uniform noise between 0 and 1 to each pixel value. We normalized the NORB dataset by $256$ and the SVHN dataset by the standard deviation on each color channel. Both datasets were assumed Gaussian distributed for the generative models $p_\theta(x|\cdot)$.

\begin{figure*}
\centering
	\begin{subfigure}{.22\textwidth}
	  \centering
      \vspace{5mm}
      \includegraphics[scale=0.55]{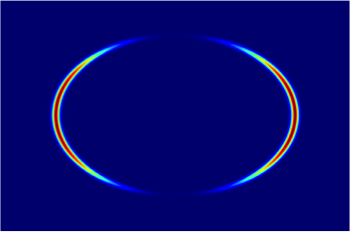}
      \vspace{8mm}
      \caption{ }
      \label{fig:posterior}
	\end{subfigure}
    \begin{subfigure}{.22\textwidth}
	  \centering
      \vspace{5mm}
      \includegraphics[scale=0.55]{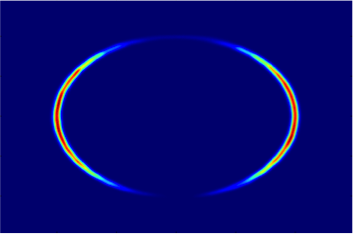}
      \vspace{8mm}
      \caption{ }
      \label{fig:fit}
	\end{subfigure}
    \begin{subfigure}{.22\textwidth}
	  \centering
      \includegraphics[scale=0.5]{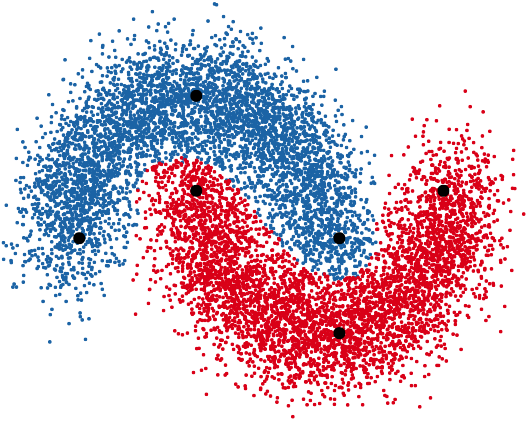}
      \vspace{-5mm}
      \caption{ }
	  \label{fig:halfmoon}
	\end{subfigure}
    \begin{subfigure}{.22\textwidth}
	  \centering
      \includegraphics[scale=0.45]{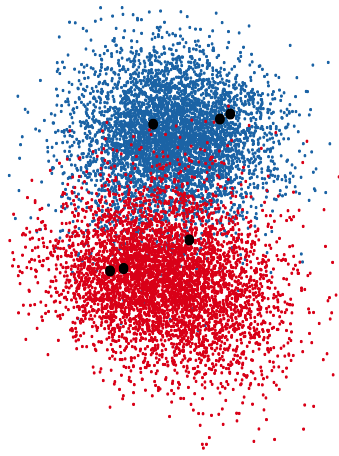}
	  \caption{ }
      \label{fig:aux}      
	\end{subfigure}
    \vspace{-3mm}
	\caption{(a) True posterior of the prior $p(z)$. (b) The approximation $q_\phi(z|a)q_\phi(a)$ of the ADGM. (c) Prediction on the half-moon data set after 10 epochs with only 3 labeled data points (black) for each class. (d) PCA plot on the 1st and 2nd principal component of the corresponding auxiliary latent space.}
\vspace{-2mm}
\end{figure*}

\section{Results} 
In this section we present two toy examples that shed light on how the auxiliary variables improve the distribution fit. Thereafter we investigate the unsupervised generative log-likelihood performance followed by semi-supervised classification performance on several benchmark datasets. We demonstrate state-of-the-art performance and show that adding auxiliary variables increase both classification performance and convergence speed (cf. Sec. \ref{sec:experimenta_details} for details).

\subsection{Beyond Gaussian latent distributions}\label{sec:vae_toy}%
In variational auto-encoders the inference model $q_\phi(z|x)$ is parameterized as a fully factorized Gaussian. We demonstrate that the auxiliary model can fit complicated posterior distributions for the latent space. To do this we consider the 2D potential model $p(z)=\exp(U(z))/Z$ \citep{Rezende2015} that leads to the bound
\begin{align}
\log Z \ge \mathbb{E}_{q_\phi(a,z)}
\left[ \log
\frac{\exp(U(z)) p_\theta(a|z)}{  q_\phi(a)q_\phi(z|a)} 
\right] .
\end{align}
Fig. \ref{fig:posterior} shows the true posterior and Fig. \ref{fig:fit} shows a density plot of $z$ samples from $a\sim q_\phi(a)$ and $z\sim q_\phi(z|a)$ from a trained ADGM. This is similar to the findings of \citet{Rezende2015} in which they demonstrate that by using normalizing flows they can fit complicated posterior distributions. The most frequent solution found in optimization is not the one shown, but one where $Q$ fits only one of the two equivalent modes. The one and two mode solution will have identical values of the bound so it is to be expected that the simpler single mode solution will be easier to infer.

\subsection{Semi-supervised learning on two half-moons}\label{sec:halfmoons}
To exemplify the power of the ADGM for semi-supervised learning we have generated a 2D synthetic dataset consisting of two half-moons (top and bottom), where $(x_\text{top}, y_\text{top})=$ $(cos([0,\pi]),sin([0,\pi]))$ and $(x_\text{bottom}, y_\text{bottom})=$ $(1-cos([0,\pi]),1 - sin([0,\pi])-0.5)$, with added Gaussian noise. The training set contains 1e4 samples divided into batches of 100 with $3$ labeled data points in each class and the test set contains 1e4 samples. A good semi-supervised model will be able to learn the data manifold for each of the half-moons and use this together with the limited labeled information to build the classifier. 

The ADGM converges close to $0\%$ classification error in 10 epochs (cf. Fig. \ref{fig:halfmoon}), which is much faster than an equivalent model without the auxiliary variable that converges in more than $100$ epochs. When investigating the auxiliary variable we see that it finds a discriminating internal representation of the data manifold and thereby aids the classifier (cf. Fig. \ref{fig:aux}).

\subsection{Generative log-likelihood performance}\label{sec:loglikelihood}
We evaluate the generative performance of the unsupervised auxiliary model, AVAE, using the MNIST dataset. The inference and generative models are defined as
\begin{align}
q_{\phi}(a,z|x)&=q_{\phi}(a_1|x)q_{\phi}(z_1|a_1,x) \\
&\quad\quad\prod_{i=2}^{L} q_\phi(a_i|a_{i-1},x) q_\phi(z_i|a_i,z_{i-1})\ , \nonumber \\
p_\theta(x,a,z)&=p_\theta(x|z_1)p(z_L)p_\theta(a_L|z_L)\\
&\quad\quad\prod_{i=1}^{L-1}p_\theta(z_i|z_{i+1})p_\theta(a_i|z_{\geq i}) \ . \nonumber
\end{align}
where $L$ denotes the number of stochastic layers. 

We report the lower bound from Eq. (\ref{eq:avae_bound}) for 5000 importance weighted samples and use the same training and parameter settings as in \citet{Sonderby2016} with warm-up\footnote{Temperature on the KL-divergence going from 0 to 1 within the first 200 epochs of training.}, batch normalization and 1 Monte Carlo and IW sample for training.

\begin{table}[!h]
\begin{center}
\begin{small}
\begin{sc}
\resizebox{\columnwidth}{!}{%
\begin{tabular}{l|c}
\hline
\abovespace
  & $\leq \log p(x)$ \\
\hline
\abovespace 
VAE+NF, L=1 \citep{Rezende2015} & $-85.10$ \\ 
IWAE, L=1, IW=1 \citep{Burda15} & $-86.76$ \\
IWAE, L=1, IW=50 \citep{Burda15} & $-84.78$\\
IWAE, L=2, IW=1 \citep{Burda15} & $-85.33$ \\
IWAE, L=2, IW=50 \citep{Burda15} & $-82.90$\\
VAE+VGP, L=2 \citep{Tran2015} & $-81.90$\\
LVAE, L=5, IW=1 \citep{Sonderby2016} & $-82.12$  \\
LVAE, L=5, FT, IW=10 \citep{Sonderby2016} & $-81.74$\\
 \hline
 \abovespace 
 Auxiliary VAE (AVAE), L=1, IW=1 & $-84.59$ \\
 Auxiliary VAE (AVAE), L=2, IW=1 & $-82.97$ \\
\hline
\end{tabular}%
}
\end{sc}
\end{small}
\end{center}
\vskip -0.1in
\caption{Unsupervised test log-likelihood on permutation invariant MNIST for the normalizing flows VAE (VAE+NF), importance weighted auto-encoder (IWAE), variational Gaussian process VAE (VAE+VGP) and Ladder VAE (LVAE) with FT denoting the finetuning procedure from \citet{Sonderby2016}, IW the importance weighted samples during training, and L the number of stochastic latent layers $z_1,..,z_L$.} \label{table:unsupervised_benchmarks}
\vspace{-2mm}
\end{table}

\begin{table*}
\begin{center}
\begin{small}
\begin{sc}
\begin{tabular}{l|lll}
\hline
\abovespace\
  & MNIST   & SVHN & NORB \\   
  & $100$ labels & $1000$ labels & $1000$ labels \\
\hline
\abovespace 
M1+TSVM \citep{Kingma14}& $11.82$\% ($\pm 0.25$) & $55.33$\% ($\pm 0.11$) & $18.79$\% ($\pm 0.05$)  \\ 
M1+M2 \citep{Kingma14} & $3.33$\% ($\pm 0.14$) &$36.02$\%($\pm0.10$) & - \\
VAT \citep{Miyato15} & $2.12$\% & $24.63$\% & $9.88$\% \\
Ladder Network \citep{Rasmus15}& $1.06$\% ($\pm 0.37$) & - & - \\ 
 \hline
 \abovespace 
 Auxiliary Deep Generative Model (ADGM) & \textbf{0.96}\% ($\pm 0.02$) & $22.86$\% & $10.06$\% ($\pm 0.05$) \\
 Skip Deep Generative Model (SDGM) & $1.32$\% ($\pm 0.07$) & \textbf{16.61}\% ($\pm 0.24$) & \textbf{9.40}\% ($\pm 0.04$) \\
\hline
\end{tabular}%
\end{sc}
\end{small}
\end{center}
\vskip -0.1in
\caption{Semi-supervised test error \% benchmarks on MNIST, SVHN and NORB for randomly labeled and evenly distributed data points. The lower section demonstrates the benchmarks of the contribution of this article.}\label{table:benchmarks}
\vspace{-2mm}
\end{table*}

We evaluate the negative log-likelihood for the 1 and 2 layered AVAE. We found that warm-up was crucial for activation of the auxiliary variables. Table \ref{table:unsupervised_benchmarks} shows log-likelihood scores for the permutation invariant MNIST dataset. The methods are not directly comparable, except for the Ladder VAE (LVAE) \citep{Sonderby2016}, since the training is performed differently. However, they give a good indication on the expressive power of the auxiliary variable model. The AVAE is performing better than the VAE with normalizing flows \citep{Rezende2015} and the importance weighted auto-encoder with 1 IW sample \citep{Burda15}. The results are also comparable to the Ladder VAE with 5 latent layers \citep{Sonderby2016} and variational Gaussian process VAE \citep{Tran2015}. As shown in \citet{Burda15} and \citet{Sonderby2016} increasing the IW samples and annealing the learning rate will likely increase the log-likelihood.

\subsection{Semi-supervised benchmarks}\label{sec:benchmarks}

\subsubsection*{MNIST experiments}
\begin{figure}
\vspace{-3mm}
	  \centering
      \includegraphics[scale=0.26]{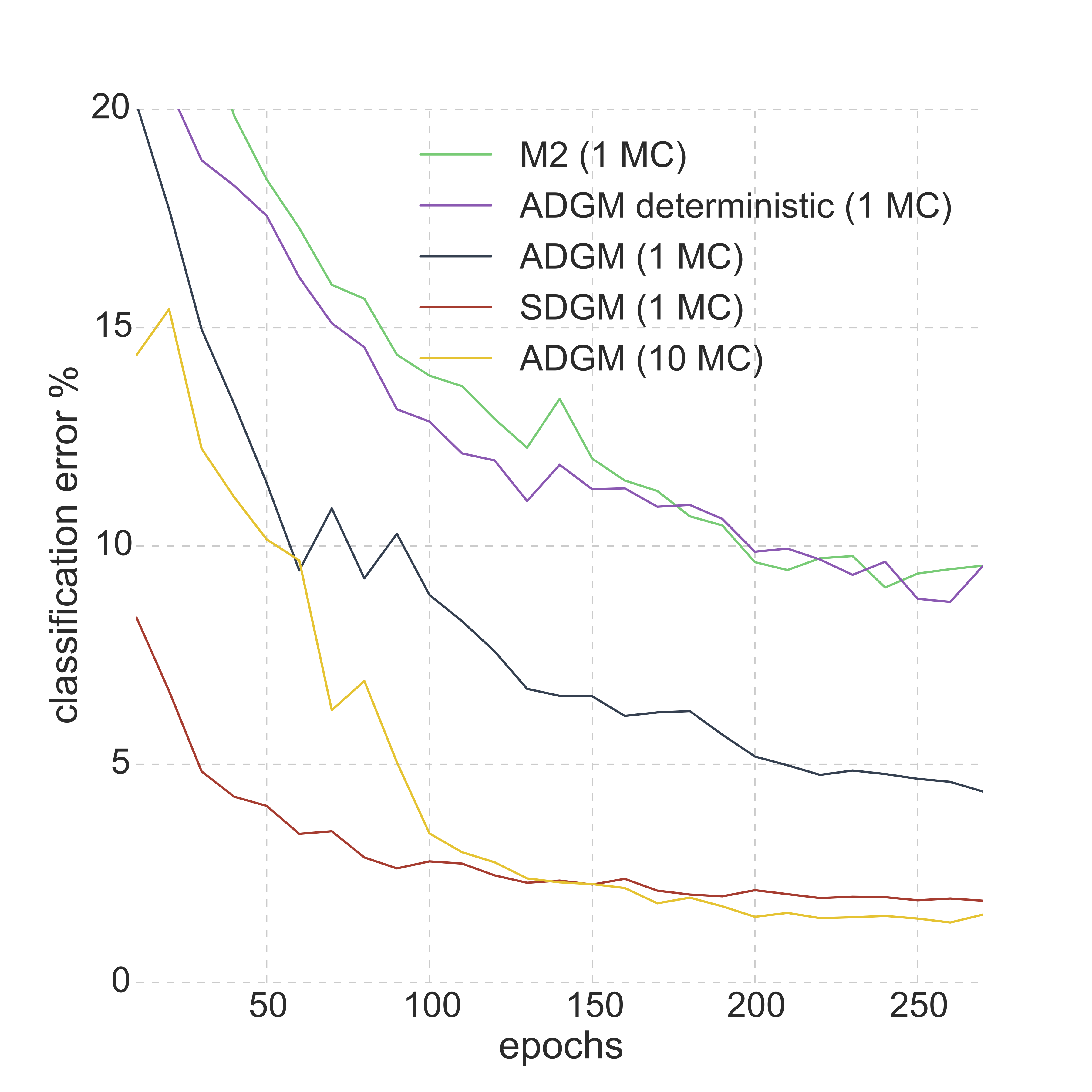}
       \vspace{-3mm}
	  \caption{100 labeled MNIST classification error $\%$ evaluated every 10 epochs between equally optimized SDGM, ADGM, M2 \citep{Kingma14} and an ADGM with a deterministic auxiliary variable.}
	  \label{fig:curve}
      \vspace{-5mm}
\end{figure}

Table \ref{table:benchmarks} shows the performance of the ADGM and SDGM on the MNIST dataset. The ADGM's convergence to around 2\% is fast (around 200 epochs), and from that point the convergence speed declines and finally reaching $0.96$\% (cf. Fig. \ref{fig:curve}). The SDGM shows close to similar performance and proves more stable by speeding up convergence, due to its more advanced generative model. We achieved the best results on MNIST by performing multiple Monte Carlo samples for $a\sim q_\phi(a|x)$ and $z\sim q_\phi(z|a,y,x)$.

\begin{figure}[h!]
  \centering
  \begin{subfigure}{.40\textwidth}
  \centering
  \includegraphics[scale=0.4]{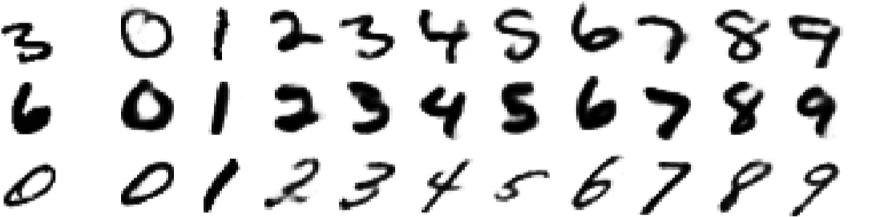}
  \caption{ }
  \label{fig:mnist_top}
  \end{subfigure}
  \begin{subfigure}{.40\textwidth}
  \centering
  \includegraphics[scale=0.4]{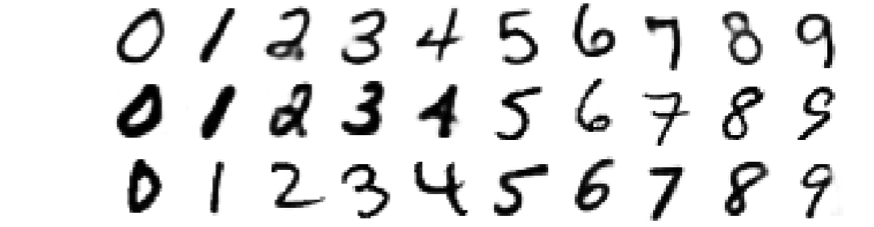}
  \caption{ }
  \label{fig:mnist_bottom}
  \end{subfigure}
   \vspace{-2mm}
  \caption{MNIST analogies. (a) Forward propagating a data point $x$ (left) through $q_\phi(z|a,x)$ and generate samples $p_\theta(x|y_j,z)$ for each class label $y_j$ (right). (b) Generating a sample for each class label from random generated Gaussian noise; hence with no use of the inference model.}
  \label{fig:mnist}
  \vspace{-4mm}
\end{figure}

A good explorative estimate of the models ability to comprehend the data manifold, or in other words be as close to the posterior distribution as possible, is to evaluate the generative model. In Fig. \ref{fig:mnist_top} we show how the SDGM, trained on \emph{only} 100 labeled data points, has learned to separate style and class information. Fig \ref{fig:mnist_bottom} shows random samples from the generative model.

\begin{figure}[h!]
\centering
\begin{subfigure}{.4\textwidth}
\centering
\includegraphics[scale=0.55]{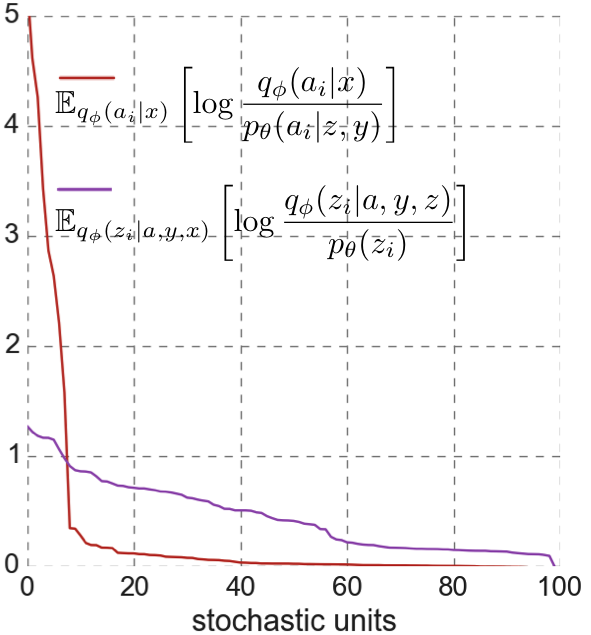}
\caption{ }
\label{fig:kl}
\end{subfigure}
\begin{subfigure}{.4\textwidth}
\centering
\includegraphics[width=.7\textwidth]{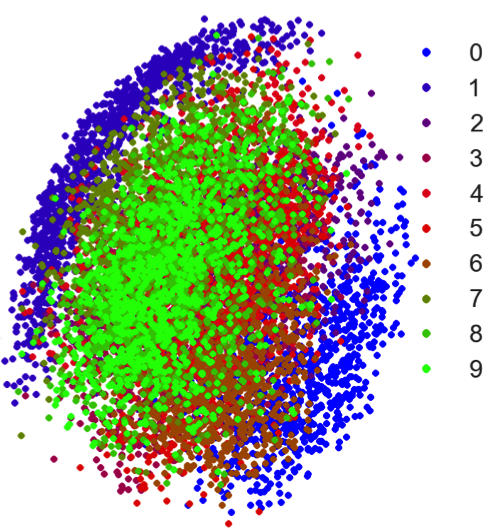}
\caption{ }
\label{fig:pca}
\end{subfigure}
 \vspace{-2mm}
\caption{SDGM trained on 100 labeled MNIST. (a) The KL-divergence for units in the latent variables $a$ and $z$ calculated by the difference between the approximated value and its prior. (b) PCA on the 1st and 2nd principal component of the auxiliary latent space.}
\end{figure}

Fig. \ref{fig:kl} demonstrate the information contribution from the stochastic unit $a_i$ and $z_j$ (subscripts $i$ and $j$ denotes a unit) in the SDGM as measured by the average over the test set of the KL-divergence between the variational distribution and the prior. Units with little information content will be close to the prior distribution and the KL-divergence term will thus be close to 0. The number of clearly activated units in $z$ and $a$ is quite low $\sim 20$, but there is a tail of slightly active units, similar results have been reported by \citet{Burda15}. It is still evident that we have information flowing through both variables though. Fig. \ref{fig:aux} and \ref{fig:pca} shows clustering in the auxiliary space for both the ADGM and SDGM respectively.

In order to investigate whether the stochasticity of the auxiliary variable $a$ or the network depth is essential to the models performance, we constructed an ADGM with a deterministic auxiliary variable. Furthermore we also implemented the M2 model of \citet{Kingma14} using the exact same hyperparameters as for learning the ADGM. Fig.~\ref{fig:curve} shows how the ADGM outperforms both the M2 model and the ADGM with deterministic auxiliary variables. We found that the convergence of the M2 model was highly unstable; the one shown is the best obtained.

\subsubsection*{SVHN \& NORB experiments}
From Table \ref{table:benchmarks} we see how the SDGM outperforms VAT with a relative reduction in error rate of more than $30$\% on the SVHN dataset. We also tested the model performance, when we omitted the SVHN \textit{extra} set from training. Here we achieved a classification error of $29.82$\%. The improvements on the NORB dataset was not as significant as for SVHN with the ADGM being slightly worse than VAT and the SDGM being slightly better than VAT.

On SVHN the model trains to around $19$\% classification error in 100 epochs followed by a decline in convergence speed. The NORB dataset is a significantly smaller dataset and the SDGM converges to around $12$\% in 100 epochs. We also trained the NORB dataset on single images as opposed to image pairs (half the dataset) and achieved a classification error around $13$\% in 100 epochs.

For Gaussian input distributions, like the image data of SVHN and NORB, we found the SDGM to be more stable than the ADGM.

\section{Discussion}
The ADGM and SDGM are powerful deep generative models with relatively simple neural network architectures. They are trainable end-to-end and since they follow the principles of variational inference there are multiple improvements to consider for optimizing the models like using the importance weighted bound or adding more layers of stochastic variables. Furthermore we have only proposed the models using a Gaussian latent distribution, but the model can easily be extended to other distributions \citep{Ranganath2014, Ranganath2015}.

One way of approaching the stability issues of the ADGM, when training on Gaussian input distributions $x$ is to add a \emph{temperature} weighting between discriminative and stochastic learning on the KL-divergence for $a$ and $z$ when estimating the variational lower bound \cite{Sonderby2016}. We find similar problems for the Gaussian input distributions in \citet{Oord2015}, where they restrict the dataset to ordinal values in order to apply a softmax function for the output of the generative model $p(x|\cdot)$. This discretization of data is also a possible solution. Another potential stabilizer is to add batch normalization \cite{Ioffe2015} that will ensure normalization of each output batch of a fully connected hidden layer. 

A downside to the semi-supervised variational framework is that we are summing over all classes in order to evaluate the variational bound for unlabeled data. This is a computationally costly operation when the number of classes grow. In this sense, the Ladder network has an advantage. A possible extension is to sample $y$ when calculating the unlabeled lower bound $-\mathcal{U}(x_u)$, but this may result in gradients with high variance.

The framework is implemented with fully connected layers. VAEs have proven to work well with convolutional layers so this could be a promising step to further improve the performance. Finally, since we expect that the variational bound found by the auxiliary variable method is quite tight, it could be of interest to see whether the bound for $p(x,y)$ may be used for classification in the Bayes classifier manner $p(y|x) \propto p(x,y)$.
\vspace{-2mm}
\section{Conclusion}
We have introduced a novel framework for making the variational distributions used in deep generative models more expressive. In two toy examples and the benchmarks we investigated how the framework uses the auxiliary variables to learn better variational approximations. Finally we have demonstrated that the framework gives state-of-the-art performance in a number of semi-supervised benchmarks and is trainable end-to-end. 

\appendix 
\section{Auxiliary model specification} \label{app:auxiliary_variable}

In this appendix we study the theoretical optimum of the auxiliary variational bound found by functional derivatives of the variational objective. In practice we will resort to restricted deep network parameterized distributions. But this analysis nevertheless shed some light on the properties of the optimum. Without loss of generality we consider only auxiliary $a$ and latent $z$: $p(a,z)=p(z)p(a|z)$, $p(z)=f(z)/Z$ and $q(a,z)=q(z|a)q(a)$. The results can be extended to the full semi-supervised setting without changing the overall conclusion. The variational bound for the auxiliary model is
\begin{align}
\log Z & \ge \mathbb{E}_{q(a,z)} \left [ \log \frac{f(z)p(a|z)}{q(z|a)q(a)} \right ] \ . 
\end{align}
We can now take the functional derivative of the bound with respect to $p(a|z)$. This gives the optimum $p(a|z)=q(a,z)/q(z)$, which in general is intractable because it requires marginalization: $q(z)=\int q(z|a)q(a) da$. 

One may also restrict the generative model to an uninformed $a$-model: $p(a,z)=p(z)p(a)$. Optimizing with respect to $p(a)$ we find $p(a)=q(a)$. When we insert this solution into the variational bound we get
\begin{align}
\int q(a) \, \mathbb{E}_{q(z|a)} \left [ \log \frac{f(z)}{q(z|a)} \right ] da \ .
\end{align}
The solution to the optimization with respect to $q(a)$ will simply be a $\delta$-function at the value of $a$ that optimizes the variational bound for the $z$-model. So we fall back to a model for $z$ without the auxiliary as also noted by \citet{Agakov04}. 

We have tested the uninformed auxiliary model in semi-supervised learning for the benchmarks and we got competitive results for MNIST but not for the two other benchmarks. We attribute this to two factors: in semi-supervised learning we add an additional classification cost so that the generic form of the objective is 
\begin{align}
\log Z & \ge \mathbb{E}_{q(a,z)} \left [ \log \frac{f(z)p(a)}{q(z|a)q(a)} + g(a) \right ] \ , 
\end{align}
we keep $p(a)$ fixed to a zero mean unit variance Gaussian and we use deep iid models for $f(z)$, $q(z|a)$ and $q(a)$. This taken together can lead to at least a local optimum which is different from the collapse to the pure $z$-model. 

\newpage
\section{Variational bounds} \label{app:variational_bounds}
In this appendix we give an overview of the variational objectives used.
The generative model $p_\theta(x,a,y,z)$ for the \emph{Auxiliary Deep Generative Model} and the \emph{Skip Deep Generative Model} are defined as
\begin{align}
&\text{ADGM:}\nonumber\\  
&p_\theta(x,a,y,z) = p_\theta(x|y,z)p_\theta(a|x,y,z)p(y)p(z)\ . \\
\nonumber\\
&\text{SDGM:}\nonumber\\  
&p_\theta(x,a,y,z) = p_\theta(x|a,y,z)p_\theta(a|x,y,z)p(y)p(z)\ .
\end{align}
The lower bound $-\mathcal{L}(x,y)$ on the labeled log-likelihood is defined as
\begin{align}
\log p(x,y) &= \log \int_a \int_z p_\theta(x,y,a,z) dzda\\
&\ge \mathbb{E}_{q_\phi(a,z|x,y)} \left [\log \frac{p_\theta(x,y,a,z)}{q_\phi(a,z|x,y)} \right] \equiv -\mathcal{L}(x,y)\ , \nonumber
\end{align}
where $q_\phi(a,z|x,y)=q_\phi(a|x)q_\phi(z|a,y,x)$. We define the function $f(\cdot)$ to be $f(x,y,a,z) = \log \frac{p_\theta(x,y,a,z)}{q_\phi(a,z|x,y)}$. In the lower bound for the unlabeled data $-\mathcal{U}(x)$ we treat the discrete $y$\footnote{$y$ is assumed to be multinomial but the model can easily be extended to different distributions.} as a latent variable. We rewrite the lower bound in the form of \citet{Kingma14}:
\begin{align}
\log p(x) &= \log \int_a \sum_y \int_z p_\theta(x,y,a,z) dzda \nonumber\\
&\ge \mathbb{E}_{q_\phi(a,y,z|x)} \left [ f(\cdot) - \log q_\phi(y|a,x) \right] \\
&= \mathbb{E}_{q_\phi(a|x)}\sum_y q_\phi(y|a,x) \mathbb{E}_{q_\phi(z|a,x)} \left [ f(\cdot) \right] +  \nonumber \\
& \quad\quad \mathbb{E}_{q_\phi(a|x)} \big[\underbrace{-\sum_y q_\phi(y|a,x) \log q_\phi(y|a,x)}_{\mathcal{H}(q_\phi(y|a,x))}\big] \nonumber \\
&= \mathbb{E}_{q_\phi(a|x)} \bigg[\sum_y q_\phi(y|a,x) \mathbb{E}_{q_\phi(z|a,x)}\left[f(\cdot)\right] + \nonumber\\
&\quad\quad\quad\quad\quad\quad\quad\quad\quad\quad\quad\quad\ \  \mathcal{H}(q_\phi(y|a,x)) \bigg] \nonumber \\ 
&\equiv -\mathcal{U}(x)\ ,\nonumber
\end{align}
where $\mathcal{H}(\cdot)$ denotes the entropy. The objective function of $-\mathcal{L}(x,y)$ and $-\mathcal{U}(x)$ are given in Eq. (\ref{eq:elbo_trick}) and Eq. (\ref{eq:elbo_col}).

\newpage
\section*{Acknowledgements} 
We thank Durk P.\ Kingma and Shakir Mohamed for helpful discussions. This research was supported by the Novo Nordisk Foundation, Danish Innovation Foundation and the NVIDIA Corporation with the donation of TITAN X and Tesla K40 GPUs.

\bibliography{references}
\bibliographystyle{icml2016}

\end{document}